\documentclass[reprint,amsmath,amssymb,aps,showkeys,longbibliography]{revtex4-2}

\usepackage{slashed}
\usepackage{amsmath}

\usepackage{graphicx}
\usepackage{dcolumn}
\usepackage{bm}
\usepackage{hyperref}
\usepackage[mathlines]{lineno}
\usepackage{xcolor}
\usepackage[makeroom]{cancel}
\usepackage{upgreek}

\usepackage{mathrsfs}

\newcommand\RR{\mathbb{R}}

\renewcommand\H{\mathcal{H}}

\newcommand\rarrow{\rightarrow}

\begin{document}


\title{Toward Manifest Relationality in Transformers via Symmetry Reduction} 

\author{J. François}
\email{jordan.francois@uni-graz.at}
\affiliation{University of Graz (Uni Graz), 
Heinrichstraße 26/5, 8010 Graz, Austria, and \\
Masaryk University (MUNI), 
Kotlářská 267/2, Veveří, Brno, Czech Republic, and \\
Mons University (UMONS), 
20 Place du Parc, 7000 Mons, Belgium. 
}

\author{L. Ravera}
\email{lucrezia.ravera@polito.it}
\affiliation{Politecnico di Torino (PoliTo),
C.so Duca degli Abruzzi 24, 10129 Torino, Italy, and \\
Istituto Nazionale di Fisica Nucleare (INFN), Section of Torino,
Via P. Giuria 1, 10125 Torino, Italy, and \\
Grupo de Investigación en Física Teórica,
Universidad Cat\'{o}lica De La Sant\'{i}sima Concepci\'{o}n, Alonso de Ribera 2850, Concepción, Chile.
}

\date{\today}

\begin{abstract}
Transformer models contain substantial internal redundancy arising from coordinate-dependent representations and continuous symmetries, in model space and in head space, respectively. 
While recent approaches address this by explicitly breaking symmetry, we propose a complementary framework based on symmetry reduction. 
We reformulate representations, attention mechanisms, and optimization dynamics in terms of invariant relational quantities, eliminating redundant degrees of freedom by construction. 
This perspective yields architectures that operate directly on relational structures, providing a principled geometric framework for reducing parameter redundancy and analyzing optimization.
\end{abstract}


\maketitle


\section{Introduction}\label{Introduction}  

Transformer architectures have become the dominant foundation of modern Machine Learning (ML) systems for language, vision, and multimodal reasoning \cite{devlin2018bert,brown2020gpt3,dosovitskiy2020vit,radford2021clip}. 
Since their introduction in \cite{vaswani2017attention}, transformers have replaced recurrent and convolutional models in most large-scale sequence modeling tasks.

At a high level, a transformer processes sequences by representing each token as a vector in a shared hidden feature space, $x_i \in \RR^d$, where $d$ is the model (hidden) dimension (all layers act on and update these vectors within the same ambient space $\RR^d$), and repeatedly refining these vectors through \emph{attention mechanisms} and \emph{feedforward layers}. 
Attention allows each token representation to update itself by referencing all other tokens (unless causal masking is applied) in the sequence, enabling flexible modeling of long-range dependencies and contextual relationships. 
Stacking such layers yields deep models capable of learning complex linguistic and semantic structures.

Despite their empirical success, transformers remain poorly understood from a theoretical standpoint \cite{merrill2020formal,zhou2023survey,elhage2023circuits}. 
Recent research has begun to analyze symmetry and equivariance in deep learning \cite{cohen2016gcnn,kondor2018generalization}, as well as optimization dynamics and representational geometry \cite{kunin2020neuralmechanics,tanaka2021noether}.
{Here, let us remark that, in transformers, equivariance means that if you apply a transformation $g$ to the input (or to an internal representation) and there is a corresponding transformation 
$\rho(g)$ on the output (or next-layer representation), then the layer $F$ (here $F$ is simply the learned function implemented by the layer) commutes with that action: 
$F(g \cdot X)=\rho(g) \cdot F(X)$. 
It is an architectural symmetry constraint: the computation is designed (or happens) to respect a group action (e.g., token permutations in set/graph transformers, translations/rotations in vision, or internal head-space basis changes in attention submodules). Invariance is the special case $\rho(g)=\text{id}$, i.e. the output does not change.}

In particular, 
our fundamental focus {in this work} is the following fact:  
\emph{Transformers feature large continuous symmetries in their linear attention submodules, so that 
many parameters and representation configurations correspond to identical model behavior} \cite{dinh2017sharp,zhang2025rotationsymmetry,silverstein:2026pdf}: this large redundancy may result in non-optimal use of computational resources and/or hinder convergence.

Notably, recent works have identified continuous reparameterization symmetries in attention that go beyond the classical permutation symmetries of Multilayer Perceptrons (MLPs).
More specifically, rotation-type symmetries in self-attention have been analyzed and exploited for model fusion \cite{zhang2025rotationsymmetry},
and quotient-geometric perspectives have been used to define sharpness and geometry on symmetry-reduced manifolds, including the $GL(d_h,\RR)$ reparameterization symmetry of the value-output sector in attention heads \cite{dasilva2025hideandseek}
($d_h$ being the dimension of the head-space vector space).
On the dynamical side, symmetry can induce conserved quantities in idealized learning dynamics \cite{zhao2023symmetriesconserved}, and very recent work shows how such conserved quantities can obstruct certain efficient training schemes for transformers, motivating explicit symmetry breaking interventions \cite{silverstein:2026pdf}.
While effective, such breaking introduces preferred directions into internal representation spaces and may obscure deeper relational structure in learned representations \cite{santoro2017relationnetworks,zaheer2017deepsets,battaglia2018relational,dwivedi2020graphtransformer,ying2021graphormer}.
Our contribution is to outline a complementary symmetry reduction program: 
rather than breaking the symmetry, we formulate states, attention, and optimization directly in terms of invariant relational variables.

In this, we take inspiration from developments in 
the modern theoretical physics of
gauge field theory
which aims, rather than breaking symmetry (e.g., by selecting arbitrary coordinate frames),
to eliminate redundant degrees of freedom (d.o.f.) by working directly with \emph{manifestly invariant} variables: e.g., constraint Hamiltonian dynamics \cite{HenneauxTeitelboim1992},  symplectic reduction \cite{GuilleminSternberg1990}, or covariant phase space approaches \cite{Gieres2023}.
One tool, developed in the past decade, systematically implementing such an idea is
the \emph{Dressing Field Method} (DFM)
\cite{Francois:2017akk,FrancoisAndre:2023jmj,Francois:2024rdm,Francois:2024laf,Francois:2024xqi,Francois:2025jro,Francois:2025lqn,Francois:2025odk,Francois:2025ptj,Francois:2025shu,Francois:2025sic,Berghofer:2025ius}, which allows to extract
the \emph{relational} \cite{Rovelli:1990ph,Rovelli:2001bz,Rovelli:2013fga,Francois:2024vlr} invariant content of physical theories. 

We note here that gauge ideas already appear, e.g., in Geometric Deep Learning (GDL) when modeling data on manifolds or non-Euclidean domains (graphs, spheres, tori, etc.) -- see, e.g., \cite{gaugeDL1,theodosis2024incorporating,theodosis2024constructing,choi2025gauge,huang2025learning,strunk2025gauge,honda2026gauge}. 
In this context, `gauge' often refers to local coordinate freedom/choice of basis in tangent spaces, and gauge-equivariant convolutions that respect this freedom (for instance, no preferred frame).
These works use the geometric language of connections on (principal) fiber bundles.

In this work, we propose to adapt the DFM philosophy to \emph{transformer architectures}: Instead of modifying networks by inserting preferred coordinate directions, we consider
how transformer computations can be reformulated in terms of \emph{relational}, manifestly invariant structures, removing internal coordinate redundancies while preserving model expressivity.
We focus on components where internal redundancies appear and propose invariant reformulations:
\begin{enumerate}
    \item Dressing internal representation frames by replacing coordinate-dependent vectors with relational invariants, hence
    reformulating attention mechanisms in invariant relational form.
    \item Considering optimization dynamics in reduced, symmetry-free parameter spaces.
\end{enumerate}

Our immediate goal is not yet to replace existing architectures,  
but to establish a conceptual and mathematical framework in which learning proceeds directly on meaningful relational d.o.f. rather than on arbitrary coordinate representations. 
Such a formulation may reduce optimization degeneracies, clarify internal mechanisms, and open new directions for invariant and interpretable architectures.
{In this context, let us observe that our proposal is orthogonal to the encoder-only \emph{vs} decoder-only architectural distinction: the symmetry-reduction viewpoint concerns internal representational and parameter redundancies and, in principle, may be studied in encoder, decoder, or encoder-decoder attention blocks alike.
Accordingly, we do not claim superiority over masked autoregressive decoder-only architectures in the present work.}

\begin{figure}[t]
\centering
\includegraphics[width=8.6cm]{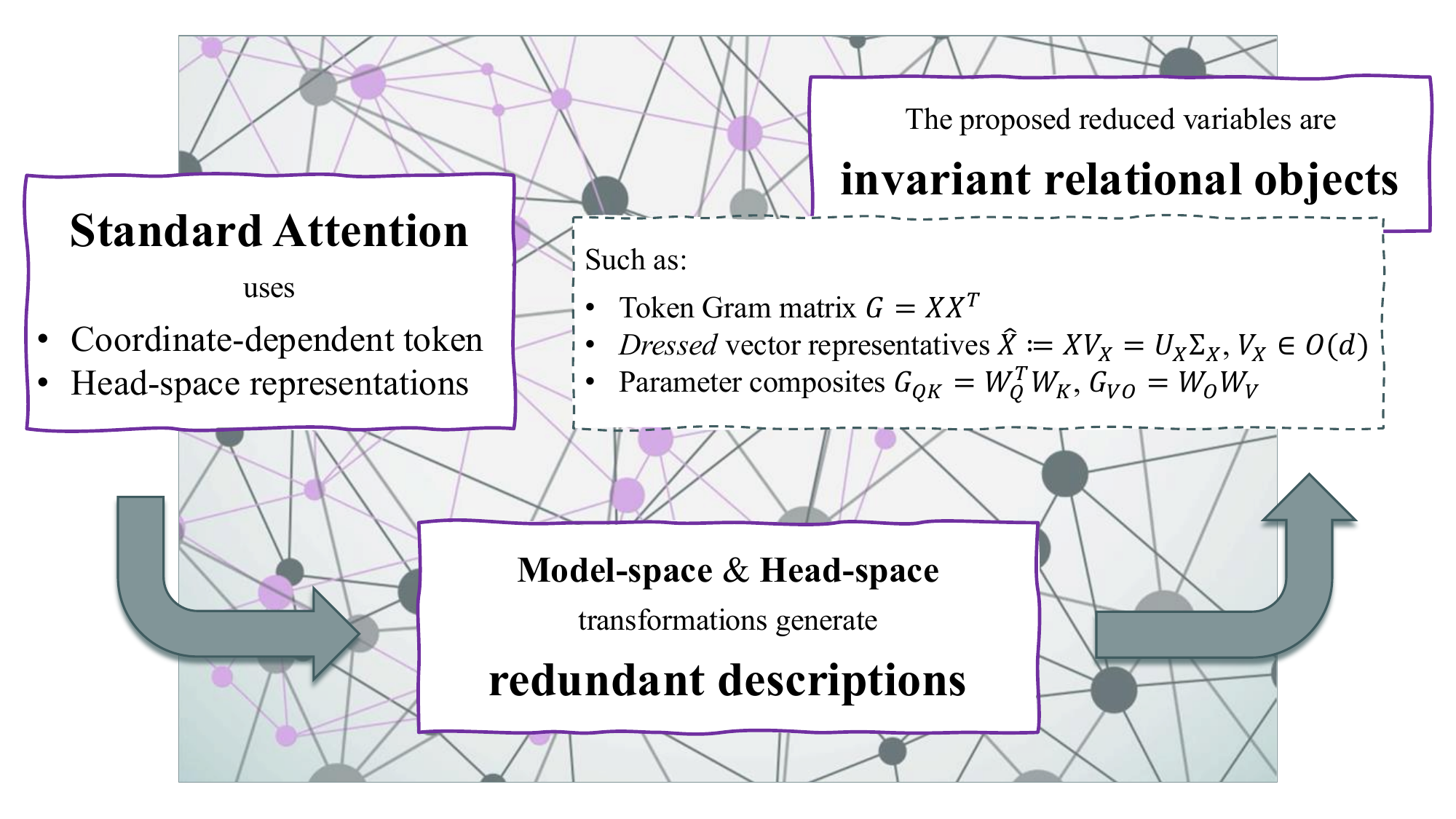}
\caption{{Schematic summary of the main content of the work.}}
\label{fig:scheme}
\end{figure}

For clarity, we distinguish three distinct (but compatible) reductions: 
(1) \emph{State/representation reduction}, which replaces coordinate-dependent hidden states $X\in\RR^{n\times d}$ ($X$ is the hidden-state matrix of one transformer layer, $n$ is the sequence length -- number of tokens, each row is one token -- and $d$ is the model/hidden dimension, or size of each token vector; so, row $i$ of $X$ is $x_i \in \RR^d$) by invariant relational objects (e.g., $XX^\top$ or $X A X^\top$, where $XX^\top \in \RR^{n\times n}$ is the \emph{Gram matrix of tokens} -- entry $(i,j)$ is $x_i^\top x_j$ -- which encodes pairwise relations between tokens and is invariant under any global rotation of the hidden space, while $XAX^\top$, with $A \in \RR^{d \times d}$, denotes a learned matrix -- entry $(i,j)$ is $x_i^\top A x_j$ -- or learned relational kernel, a more flexible invariant similarity than plain dot products); 
(2) \emph{Parameter reduction}, which replaces redundant parameterizations by invariant composites (e.g., $W_Q^\top W_K$, with $W_Q$ the \emph{query} projection matrix and $W_K$ the \emph{key} projection matrix, both $\in \RR^{d_h \times d}$ -- they are both learned linear maps{, they take a token vector 
$x_i \in \mathbb{R}^d$
and produce
$q_i = W_Q x_i$, 
$k_i = W_K x_i$,
where 
$q_i,\, k_i \in \mathbb{R}^{d_h}$
live in the head space;
then, attention scores are
$q_i^\top k_j$,
which is just the dot product between query and key}) that determine the forward computation;
(3) \emph{Dynamical/optimization reduction}, which modifies training so updates have no component along symmetry orbits (quotient or projected dynamics). 

Sections \ref{Internal representation frames in Machine Learning} and \ref{Dressing and relational invariants in model space} of this paper focus on (1), while Section \ref{Optimization dynamics on reduced, symmetry-free parameter spaces} is devoted to (2)-(3).
{Fig. \ref{fig:scheme} provides a schematic overview of our symmetry-reduction perspective.
A complementary summary of the redundancies and the corresponding invariant variables is given in Table \ref{tab:reductions_main}.}

\begin{table}[h!]
\caption{\label{tab:reductions_main}
{Redundancies and corresponding invariants.}}
\begin{ruledtabular}
\begin{tabular}{ll}
\textbf{Redundancy type} & \textbf{Invariant variable} \\
Model-space frame
& $G=XX^\top$, $XAX^\top$, $\hat X=Xu[X]^{-1}$ \\
QK head-space symm. 
& $G_{QK}=W_Q^\top W_K$ \\
VO head-space symm. 
& $G_{VO}=W_O W_V$ \\
\end{tabular}
\end{ruledtabular}
\end{table}

Throughout this paper we distinguish two independent sources of redundancy in transformer architectures and address them separately:
\begin{enumerate}
\item[(I)] \emph{Model space (representation-frame) redundancy}, corresponding to arbitrary orthogonal changes of basis
$X\mapsto XU^\top$, $U\in O(d)$ in the model feature space $\mathbb R^d$.
This motivates a reformulation of hidden states and attention weights in terms of invariant relational quantities (Sections \ref{Internal representation frames in Machine Learning} and \ref{Dressing and relational invariants in model space}). \\
\emph{Part I. Relational Representation and Model Space Symmetry Reduction in $\RR^d$} of the manuscript is devoted to this.

\item[(II)] \emph{Parameter space reparameterization redundancy} inside attention heads, corresponding to exact internal symmetries
$O(d_h)$ (query-key sector) and $GL(d_h,\mathbb R)$ (value-output sector).
These symmetries act on parameters while leaving the computed function unchanged and are treated via symmetry-reduced optimization and invariant composites (Section \ref{Optimization dynamics on reduced, symmetry-free parameter spaces}). \\
\emph{Part II. Attention-Head Reparameterization Symmetry and Optimization Dynamics in $\RR^{d_h}$} of the paper concerns this.
\end{enumerate}
The first reduction concerns \emph{representations}; the second concerns \emph{parameters}.
They are logically independent but complementary.

The remainder of this paper is structured as follows. In Section \ref{The Dressing Field Method in field theory: Conceptual motivation}, we briefly review the DFM in its field-theoretic setting, emphasizing its role as conceptual motivation for symmetry reduction via relational variables. 
Section \ref{Internal representation frames in Machine Learning} discusses internal representation frames in transformer models. 
In Section \ref{Dressing and relational invariants in model space}, we reformulate token representations in model space in relational terms via ``dressing". 
Section \ref{Optimization dynamics on reduced, symmetry-free parameter spaces} addresses optimization dynamics on reduced, symmetry-free parameter spaces, hence in head space. 
{There, we also provide a concrete differentiable realization and initialization sketch
for a representative case, that is the value-output sector of a single attention head.}
Section \ref{Conclusions} concludes the paper with directions for future work{, including a concrete evaluation roadmap}.
In Appendix \ref{Algorithmic sketch: Invariant updates for the value-output sector}, we provide an illustrative algorithmic sketch of symmetry-reduced optimization for a representative attention submodule.

Let us remark that this paper develops a conceptual and mathematical framework.
We (i) identify precise reparameterization symmetries for idealized transformer blocks, (ii) show how invariant composites encode the functionally relevant d.o.f. in specific submodules (e.g., query-key -- Q/K -- bilinear forms), and (iii) suggest
symmetry-reduced architectural and optimization realizations.
We do not claim that every practical transformer with LayerNorm (\emph{Layer Normalization}, which normalizes each token vector independently across its feature dimension) and standard implementation details enjoys the same full symmetry group; 
rather, we treat such details as reducing the symmetry and motivating approximate or partial reductions.
Empirical evaluation and engineering trade-offs are left for future work.

\paragraph{{Scope of the present work.}\\}

{
The purpose of this paper is conceptual and mathematical.
We develop a symmetry-reduction framework for transformer representations and optimization, identify the relevant invariant composites in idealized attention submodules, and formulate corresponding relational variables.
We also provide concrete differentiable realizations for representative sectors (see Section \ref{Concrete realization and initialization sketch}), illustrating how invariant variables, gradients, and initialization may be handled in practice.}

{We do not, however, present a benchmarked implementation or claim empirical speedups, accuracy gains, or reduced training cost on standard datasets.
Such questions require a dedicated engineering and experimental study, which lies beyond the scope of the present article.
To make the empirical implications of the framework precise, we provide in Section \ref{Empirical scope and evaluation roadmap} a concrete evaluation roadmap for future work.}

\section{The Dressing Field Method in field theory: Conceptual motivation}
\label{The Dressing Field Method in field theory: Conceptual motivation}

The purpose of this section is conceptual rather than technical. We do not claim that transformer models possess gauge symmetries in the field-theoretic sense, nor that the Dressing Field Method (DFM) can be directly applied to ML architectures. 
Instead, we briefly review the DFM in its native theoretical physics context as a mature framework whose central insight -- the elimination of redundant d.o.f. through invariant, relational variables -- serves as methodological \emph{inspiration} for the symmetry-reduction program suggested here.

In Gauge Field Theory (GFT), the DFM provides a systematic procedure for replacing gauge-dependent fields by relational, gauge-invariant quantities extracted from the original field content. 
Our goal is to abstract this philosophy and adapt it to transformer architectures, where internal representations exhibit analogous redundancies arising from coordinate-dependent parameterizations rather than  gauge symmetries.

We therefore present a concise overview of the DFM in the case of internal gauge symmetries, in order to motivate the relational reformulations introduced in subsequent sections.

\smallskip

In its natural theoretical physics framework, namely general-relativistic Gauge Field Theory (gRGFT), the DFM
\cite{Francois:2017akk,FrancoisAndre:2023jmj,Francois:2024rdm,Francois:2024laf,Francois:2024xqi,Francois:2025jro,Francois:2025lqn,Francois:2025odk,Francois:2025ptj,Francois:2025shu,Francois:2025sic,Berghofer:2025ius}
is used to construct gauge-invariant, relational variables \cite{Francois:2024vlr} from the field space
$\Phi=\{\upphi\}$ of a gauge theory, $\upphi$ being a collection of fields on the ``spacetime" manifold $M$.
We briefly recall the dressing procedure for internal gauge symmetry group $\mathcal{H}$, whose elements are smooth maps $\gamma: M \rarrow H$ and $H$ is a (finite-dimensional) Lie group.

\paragraph{Invariant dressed fields.\\}

Consider a gauge field theory with field content $\upphi=\{A,\varphi\}$, where $A$ denotes the gauge potential (e.g., the electromagnetic potential, when $H=U(1)$) and $\varphi$ collectively denotes matter fields. These transform under the gauge group $\mathcal{H}$ as
\begin{align}
A^\gamma := \gamma^{-1} A \gamma + \gamma^{-1} d\gamma,
\qquad
\varphi^\gamma := \gamma^{-1}\varphi. 
\label{gaugetrAvphi}
\end{align}
A dressing field is defined as a smooth map
\begin{align}
u : M \rightarrow H,
\qquad
\text{such that} \quad u^\gamma = \gamma^{-1}u,
\label{dressing-field-101}
\end{align}
with the crucial property that $u$ is extracted from the original field content, i.e. $u=u[\upphi]$. Consequently, the dressing field is an equivariant functional of the fields,
\begin{equation}
u^\gamma := u[\upphi^\gamma] = \gamma^{-1}u[\upphi].
\end{equation}
Given such a dressing field, one defines the dressed fields
\begin{align}
A^u := u^{-1}Au + u^{-1}du,
\qquad
\varphi^u := u^{-1}\varphi.
\label{dressed-fields}
\end{align}
By construction, these  are strictly $\mathcal{H}$-invariant.

When the dressing field is field-dependent, the resulting dressed variables admit a natural relational interpretation: the dressed fields $\{A^{u[A,\varphi]},\varphi^{u[A,\varphi]}\}$ encode gauge-invariant relations among the original d.o.f. in $\upphi=\{A, \varphi\}$. In this sense, the DFM replaces gauge-dependent variables with relational physical quantities.

\smallskip

The dynamics of a gauge field theory is determined by a Lagrangian $L(\upphi)$, typically required to be quasi-invariant under $\mathcal{H}$, i.e.
\begin{equation}
L(\upphi^\gamma)=L(\upphi)+db(\gamma;\upphi).
\end{equation}
This ensures the covariance of the field equation, obtained from $L(\upphi)$ via the variational principle, under $\H$. Given a dressing field $u$, one defines the dressed Lagrangian
\begin{equation}
L(\upphi^u):=L(\upphi)+db(u;\upphi),
\label{dressed-Lagrangian-int}
\end{equation}
which is strictly $\mathcal{H}$-invariant. The dressed field equations possess the same functional form as the original ones, but now govern gauge-invariant relational variables with a well-posed deterministic evolution.

\paragraph{Residual transformations.\\}

If $\mathcal{H}$ is only partially reduced, dressed fields may still transform under residual symmetries. Additionally, possible ambiguities in the choice of dressing field {are} encoded in \emph{transformations of the second kind}, 
often interpretable as physical reference frame covariance
\cite{Francois:2025shu,Francois:2025lqn}.

We emphasize that this brief review is presented solely to illustrate the general mechanism of symmetry reduction through invariant relational variables. 
In the remainder of this work, we apply this idea abstractly to transformer architectures, replacing gauge symmetry by internal representational redundancy and dressed fields/variables by relational reformulation of neural representations.

\section*{Part I. Relational Representation and Model Space Symmetry Reduction in $\RR^d$}

\section{Internal representation frames in Machine Learning}\label{Internal representation frames in Machine Learning}

A central observation motivating our proposed approach to ML is that internal representations in transformer models are not uniquely defined: 
they depend on arbitrary coordinate choices in hidden state space. 
While such choices do not affect model outputs, they introduce redundant internal d.o.f. that influence optimization dynamics and obscure the relational structure encoded in representations. 
In this section, we analyze this coordinate redundancy and propose a reformulation in which computations depend only on invariant, relational quantities rather than arbitrary coordinate frames.

\subsection{Coordinate redundancy in transformer representations}
\label{Coordinate redundancy in transformer representations}

In transformer architectures, 
hidden states at a given layer are represented as vectors
\begin{equation}
x_i \in \RR^d, \quad i = 1,\dots,n,
\end{equation}
where $n$ is the sequence length and $d$ the hidden dimension. Attention and feedforward layers act on these vectors through learned linear maps.

{For simplicity of exposition, we shall suppress positional encoding mechanisms in the main formulas.
In practical transformers, positional information may be added either additively or through structured modifications of the query-key interaction, as in rotary position embeddings (RoPE).
Such mechanisms generally constrain/modify the symmetry structure discussed here.}

Each token representation $x_i$
is linearly projected into three vectors called the query, key, and value representations:
\begin{equation}
q_i = W_Q x_i, \quad
k_i = W_K x_i, \quad
v_i = W_V x_i,
\end{equation}
where the matrices
\begin{align}
W_Q, W_K, W_V \in \RR^{d_h \times d}
\end{align}
are learned parameters, and $d_h$ is the dimension of each attention head. 
Attention weights are computed from query-key similarities,
\begin{equation}
\label{mech}
\begin{aligned}
& s_{ij} = q_i^\top k_j
           = (W_Q x_i)^\top (W_K x_j), \\
& \text{Attn}(i,j) = \alpha_{ij} := \text{softmax}_j\left(\frac{s_{ij}}{\sqrt{d_h}}\right), \\
& y_i = \sum_j \alpha_{ij} v_j, 
\end{aligned}           
\end{equation}
and are then used to aggregate value vectors across tokens, producing context-dependent updates of token representations.
{We recall that the `softmax' function applied to a vector of raw scores (logits) $z = (z_1, \dots, z_n) \in \RR^n$ is defined as
$\text{softmax}(z)_i = \frac{\exp(z_i)}{\sum_{j=1}^n \exp(z_j)}$, $i=1,\dots,n$.
It maps the scores to a probability distribution (non-negative entries summing to $1$), with larger values receiving exponentially higher weight.}
Finally, we have
$W_O\in\RR^{d\times d_h}$, which projects head-space outputs in $\RR^{d_h}$ back to the model space $\RR^{d}$.
{Notice that the softmax does not break the shared $O(d_h)$ query-key symmetry because it depends only on the invariant logits $q_i^\top k_j$, but nonlinear and normalization layers outside this idealized submodule generally reduce the full symmetry.}

Although the mechanism in \eqref{mech} enables flexible contextual interactions, it remains fundamentally coordinate-based: queries, keys, and values are vectors in an internal latent space whose basis is arbitrary.

This means that transformer blocks contain \emph{reparameterization redundancies} in intermediate subspaces.

First, on the one hand,
we have a transformation acting on tokens in the \emph{model space},
\begin{equation}
\label{tokentransfright}
x_i \mapsto U x_i,
\quad\text{ with } U\in O(d),
\end{equation}
which is a change of basis in \emph{model space}.
To preserve the attention computation under this change of basis,
the weight matrices must transform as
\begin{equation}
\begin{aligned}
W_Q &\mapsto W_Q U^{-1}, \\
W_K &\mapsto W_K U^{-1}, \\
W_V &\mapsto W_V U^{-1}, \\
W_O &\mapsto U W_O .
\end{aligned}
\end{equation}
This is \emph{representation covariance}, not a parameter redundancy.

On the other hand, we have continuous symmetries arising in the \emph{linear attention submodules} as invertible changes of basis in each \emph{head space}.
Concretely, for a single head with $W_Q,W_K,W_V\in\RR^{d_h\times d}$ and $W_O\in\RR^{d\times d_h}$, the forward computation is invariant under:
\begin{align}
(W_Q,W_K) &\mapsto (S\,W_Q,\;S\,W_K), && S\in O(d_h), \label{eq:QK_Odh_sym}\\
(W_V,W_O) &\mapsto (S^{-1}W_V,\;W_O S), && S\in GL(d_h,\RR). \label{eq:VO_GLdh_sym}
\end{align}
We restrict to $O(d_h)$ since the score uses the standard Euclidean inner product in head space; a general $GL(d_h,\RR)$ transformation would not preserve, e.g., $q_i^\top k_j$ without a compensating change of metric.
Note that
(i) query-key scores depend only on head-space inner products and are preserved by the shared $O(d_h)$ rotation in \eqref{eq:QK_Odh_sym}, and (ii) the value-output contribution depends only on the composite $W_O W_V$ and is therefore invariant under the re-factorization symmetry \eqref{eq:VO_GLdh_sym}.

Consequently, attention operates in a representation space containing redundant d.o.f. 
So multiple internal parameter configurations correspond to identical model functions.
Learning dynamics may therefore evolve along directions that do not affect outputs, producing optimization degeneracies analogous to gauge redundancies in physical systems.
This motivates our proposal: 
reformulating attention directly in terms of relational invariants.

Notice, however, that a full transformer block with LayerNorm, biases, and elementwise MLP nonlinearities is not invariant under arbitrary changes of basis in the model dimension $d$; 
our relational scoring isolates an 
$O(d)$-invariant sub-computation inside this otherwise symmetry-breaking architecture (see Section \ref{What symmetry actually holds (architecture-dependent)}).

\subsubsection{What symmetry actually holds (architecture-dependent)}
\label{What symmetry actually holds (architecture-dependent)}

Before proceeding with our approach, 
to avoid overclaiming, let us fix here an explicit class of blocks and state the precise reparameterization symmetry.

\paragraph{What is symmetric.\\}

In standard transformers, most nonlinear components (MLP with elementwise activation, LayerNorm, biases) are defined with respect to a chosen coordinate system in $\RR^d$ and therefore are \emph{not} equivariant under an arbitrary $GL(d,\RR)$ change of basis in the hidden-feature dimension.
Exact continuous symmetries instead appear in specific \emph{linear submodules} (notably, \emph{attention heads}), where reparameterizations of intermediate head spaces leave the computed function unchanged (see Section \ref{Optimization dynamics on reduced, symmetry-free parameter spaces}).
Accordingly, in this paper, \emph{internal symmetry} refers to: (i) exact symmetries of such linear submodules, and (ii) approximate symmetries of more complete blocks when nonlinearities act as symmetry-breaking perturbations.

\paragraph{Effect of LayerNorm and biases.\\}

In practical transformers, LayerNorm and bias terms restrict this symmetry.
For example, standard LayerNorm is not equivariant under arbitrary $GL(d,\RR)$ basis changes.
Hence, throughout this paper, any claim of continuous internal symmetry should be interpreted as:
(i) exact for the minimal block above, and
(ii) approximate or reduced to a smaller group for common architectural choices (LayerNorm, biases, parameter tying), which we treat as symmetry-breaking perturbations of the idealized symmetry.

{This point is especially important in practice, since mainstream architectures in language and vision (e.g., BERT/GPT-type models, ViT-type models) systematically include normalization layers and nonlinear MLP blocks.
Accordingly, the exact continuous symmetries discussed here should be understood as properties of idealized linear attention submodules embedded in larger architectures where these symmetries are partially or approximately broken.
This is why we view the present work as a framework for identifying the symmetry-reduced core structure, rather than as a claim about exact full-network symmetry in production models.}

\section{Dressing and relational invariants in model space}\label{Dressing and relational invariants in model space}

We start by providing our idea for a relational scoring formulation in which attention \emph{weights} are computed from the Gram matrix
\begin{equation}
\label{Gramma}
G:=XX^\top\in\RR^{n\times n},
\quad n=\text{sequence length},
\end{equation}
which is strictly invariant under $X\mapsto XU^\top$, $U\in O(d)$.
Token states are represented relationally by the Gram data, e.g., by the $i$-th row $(G_{ij})_{j=1}^n$ which encodes the relations of token $i$ to all other tokens in the current sequence, rather than by coordinate-dependent vectors $x_i\in\RR^d$.
When convenient, we still use $X$ as an $O(d)$-equivariant representative to transport vector-valued features, while the invariant content used for scoring is carried by $G$.

We present this $O(d)$-equivariant Gram-based attention as an idealized, symmetry-manifest alternative; incorporating generic learned feature maps (e.g., $W_V,W_O$) would generally break $O(d)$ unless explicitly constrained. Accordingly, we view the following construction as a conceptual prototype of relational attention.
We may therefore think of an attention mechanism in which attention weights depend only on
$O(d)$-invariant quantities between token representations (hence, $O(d)$-frame-invariant attention weights).
Let $X\in\RR^{n\times d}$ denote the matrix of token states (row $i$ is $x_i^\top$), and let the model feature space
transform as \eqref{tokentransfright}.
Under this action, the Gram matrix
$G := XX^\top$, $G_{ij}=x_i^\top x_j$,
is strictly invariant.

We therefore define attention scores directly from Gram invariants as 
\begin{equation}
s_{ij} = f(R_{ij})= f\!\big(G_{ij},G_{ii},G_{jj}\big),
\end{equation}
where $f:\RR^3\to\RR$ is a learned scalar function (for instance a small multilayer perceptron applied pointwise),
and $G_{ii}=\|x_i\|^2$ allows the scoring function to access token norms.
{We remark that,
when $R_{ij}$ is taken to be an inner product (or a learned bilinear form) and $f$ is scalar, the resulting attention weights are a form of \emph{kernelized attention}.
Our contribution here is not to claim novelty of pairwise-kernel attention per se, but to place such constructions in a symmetry-reduction program: (i) choose relational invariants as primary variables, and (ii) parameterize and optimize only the d.o.f. that affect these invariants.
We stress that this construction is intended as an idealized, illustrative prototype rather than a drop-in replacement for standard dot-product attention, meant to make the symmetry-reduced structure explicit.}

The attention weights are then given by
\begin{equation}
\text{Attn}(i,j)
=
\text{softmax}_j\!\left(\frac{s_{ij}}{\tau}\right),
\end{equation}
where $\tau>0$ is a temperature parameter (analogous to the conventional score normalization in dot-product attention).
By analogy with dot-product attention, one may set $\tau=\sqrt{d_h}$, but here $\tau$ is simply a temperature hyperparameter for the scalar logits $s_{ij}$.
{We restrict here to $\tau>0$, for which the softmax is smooth.
The singular limit $\tau\to 0^+$ corresponds formally to a hard argmax selection and would require a separate non-smooth treatment, beyond the scope of the present discussion.}

Since the scores depend only on Gram invariants, the attention weights are strictly $O(d)$-invariant.
To propagate vector-valued information while preserving symmetry, we take values to be the token states themselves
and define the output as
\begin{equation}
y_i=\sum_j \text{Attn}(i,j)\,x_j ,
\end{equation}
or in matrix form $Y=AX$, where $A_{ij}=\text{Attn}(i,j)$.
This update is $O(d)$-equivariant: if $X\mapsto XU^\top$, then $Y\mapsto YU^\top$.
If one wishes a \emph{fully} invariant state update, one may propagate $G$ itself:
\begin{equation}
G^{+}:=YY^\top=(AX)(AX)^\top = AGA^\top,
\end{equation}
which is strictly $O(d)$-invariant as $A$ depends only on $G$.

In this formulation, internal coordinate frames are eliminated by construction: attention operates purely on
relational invariants, and vector features are transported equivariantly.
This realizes a genuine dressed attention mechanism in which semantic interaction depends only on relations
between tokens rather than on arbitrary feature-space coordinates.

\subsection{Dressing of vector representatives}

Besides fully relational state propagation, one may realize symmetry reduction by introducing a \emph{dressing field} that selects a canonical representative of each $O(d)$-orbit while retaining vector-valued carriers.
Let $X\in\RR^{n\times d}$ denote the token matrix, transforming as
\begin{equation}
X\mapsto XU^\top,\qquad U\in O(d).
\end{equation}
A ``dressing field" is a data-dependent map
\begin{equation}
u[X]\in O(d)
\end{equation}
satisfying the equivariance property
\begin{equation}
u[XU^\top]=u[X]\,U^\top .
\end{equation}
One then defines the dressed representative
\begin{equation}
\hat X := X\,u[X]^{-1}.
\end{equation}
By construction,
\begin{equation}
X\mapsto XU^\top \;\Rightarrow\; \hat X\mapsto \hat X,
\end{equation}
so $\hat X$ is strictly $O(d)$-invariant.

A concrete choice is obtained from the singular value decomposition (SVD{, which factorizes $X$ into orthogonal matrices $U_X$, $V_X$ and a diagonal matrix $\Sigma_X$, enabling a canonical extraction of the right singular vectors $V_X$}) 
\begin{equation}
X = U_X\,\Sigma_X\,V_X^\top,
\quad V_X\in O(d),
\end{equation}
and defining
\begin{equation}
u[X]:=V_X^\top .
\end{equation}
Under $X\mapsto XU^\top$, one has (for a consistent choice of SVD) $V_{XU^\top}=U\,V_X$,
hence
\begin{equation}
u[XU^\top]=V_{XU^\top}^\top = V_X^\top U^\top = u[X]\,U^\top,
\end{equation}
as required.
The dressed representative becomes
\begin{equation}
\hat X := X V_X =U_X\,\Sigma_X.
\end{equation}
All subsequent computations may then be performed using $\hat X$, which is invariant under model space frame rotations.

\paragraph{Residual ambiguity.\\}

If $\Sigma_X$ has degeneracies or $X$ is rank-deficient, $V_X$ is not unique, leading to a residual $O(m)$ freedom 
within each $m$-dimensional degenerate singular subspace, together with discrete sign and permutation ambiguities. 
This corresponds to the standard \emph{transformations of the second kind}, familiar from DFM constructions.

Note that this realization implements symmetry reduction without eliminating vector carriers: vectors are retained for computational convenience, but internal frame redundancy is removed by dressing rather than broken.

\paragraph{Well-definedness of the dressing map.\\}

The definition $u[X]:=V_X^\top$ presupposes a deterministic convention for choosing the SVD.
Indeed, the right singular factor $V_X\in O(d)$ is not uniquely defined: even when $\Sigma_X$ has \emph{distinct} nonzero singular values, each singular vector is only defined up to an overall sign, and when $\Sigma_X$ has degeneracies (or $\text{rank}(X)<d$) there is a continuous freedom $V_X\mapsto V_X R$ with $R\in O(m)$ acting in each degenerate singular subspace.
Accordingly, the equivariance identity
$u[XU^\top]=u[X]\,U^\top$
holds on the open set where a chosen convention fixes these ambiguities (e.g., by fixing the sign of each singular vector via a deterministic rule), and otherwise holds up to the corresponding residual transformation.
In practice, one may (i) restrict the discussion to matrices $X$ with simple singular spectrum, where the map can be made locally smooth by such conventions, or (ii) treat the remaining freedom as a residual symmetry (of the second kind) and accept that $\hat X$ is defined only up to the corresponding residual action.

\medskip

Our relational formulation eliminates dependence on model space coordinate frames ($O(d)$) at the level of attention weights. 
Even after eliminating model space frame dependence, standard attention heads still possess exact internal reparameterization symmetries acting on intermediate head spaces. 
These do not affect representations directly but generate flat directions in parameter space. 
We now turn to this second, independent reduction.

In the next section, we examine how optimization dynamics themselves can be reformulated in reduced relational parameter spaces.
In fact, head-space reparameterization symmetries ($O(d_h)$, $GL(d_h)$) act in parameter space and can be removed by quotient-based optimization.
Optimization dynamics therefore evolve in reduced parameter spaces, eliminating directions that do not affect model behavior.
{Note that this symmetry reduction parallels gauge reduction in physical systems (this is an analogy, not a literal gauge symmetry): rather than introducing preferred directions, redundant d.o.f. are eliminated by construction.}
Consequently, learning trajectories are constrained to meaningful relational manifolds, potentially improving optimization efficiency and stability.

\section*{Part II. Attention-Head Reparameterization Symmetry and Optimization Dynamics in $\RR^{d_h}$}

\section{Optimization dynamics on reduced, symmetry-free parameter spaces}
\label{Optimization dynamics on reduced, symmetry-free parameter spaces}

Standard transformer parameterizations contain large internal redundancies: 
many distinct parameter settings realize the same input-output function due to continuous symmetries (e.g., internal rotations of representation subspaces) and other reparameterization freedoms in head space. 
From an \emph{optimization} perspective, such redundancies give rise to flat directions and may induce dynamics that expend update budget moving along directions that do not change the predictive behavior of the model.

In fact, when training is viewed dynamically, internal symmetries can be particularly consequential. Symmetry directions can create (approximately) conserved quantities or long-lived modes that constrain exploration and slow descent in functionally relevant directions. 
This phenomenon has been highlighted recently in Hamiltonian analyses of learning dynamics in attention models, where continuous symmetries yield conserved currents in idealized dynamics and corresponding optimization pathologies in practice \cite{silverstein:2026pdf}.

The objective of this section is to formalize a symmetry-reduction viewpoint for optimization: 
Rather than breaking symmetries by introducing preferred directions, we aim to \emph{quotient out} redundant d.o.f. and study learning directly on a reduced, symmetry-free space of functionally distinct models.

\subsection{Symmetry groups and equivalence classes in parameter space}\label{Symmetry groups and equivalence classes in parameter space}

Let $\theta \in \Theta$ denote the full parameter vector of a transformer model (all trainable weights and biases, that is, in a transformer: embedding tables, all projection matrices $W_Q$, $W_K$, $W_V$, $W_O$, MLP weights, LayerNorm parameters, etc.; $\Theta$ is the parameter space) and let
\begin{equation}
\mathcal{L}(\theta) = \mathbb{E}_{(x,y)\sim \mathcal{D}} \,\ell(f_\theta(x),y)
\end{equation}
be the population loss, that is the expected loss over the true data distribution ({which is the ideal objective one would minimize knowing $\mathcal{D}$ exactly,} with $x$ the input random variable, $y$ the target/label random variable associated with $x$, $(x,y)$ a data point, and $\mathcal{D}$ the data distribution over pairs $(x,y)$ {-- writing $(x,y)\sim \mathcal{D}$ means: `sample a random training example from the underlying population distribution'}; $\mathbb{E}_{(x,y)\sim \mathcal{D}}\,[\cdot]$ is the expectation with respect to that distribution and $\ell$ denotes the per-example loss function, or pointwise loss -- it compares the model's prediction to the true target $y$). 

Now, suppose there exists a (Lie) group $G$ acting on parameters,
\begin{equation}
\theta \mapsto g\cdot \theta,\quad g\in G,
\end{equation}
such that the model function is unchanged:
\begin{equation}
f_{g\cdot \theta}(x) = f_{\theta}(x)\quad \forall x,\ \forall g\in G.
\end{equation}
Then $\mathcal{L}(\theta)$ is constant along \emph{group orbits}:
\begin{equation}
\mathcal{L}(g\cdot \theta) = \mathcal{L}(\theta).
\end{equation}
Hence, physically (or functionally) distinct models correspond not to points in $\Theta$ but to equivalence classes
\begin{equation}
[\theta] \;:=\; \{ g\cdot \theta \,:\, g\in G\}.
\end{equation}
The \emph{reduced parameter space} is the quotient
\begin{equation}
\Theta_{\mathrm{red}} := \Theta / G,
\end{equation}
whose points correspond to distinct functions. Optimization should take place on $\Theta_{\mathrm{red}}$ rather than on $\Theta$, because motion tangent to the orbit $[\theta]$ does not change the learned function.

\subsection{Decomposition of updates: Tangent \emph{vs} transverse directions}

At a given $\theta$, the orbit direction is spanned by the infinitesimal generators of $G$. Let $\mathfrak{g}$ denote the Lie algebra of $G$ and let $\xi \in \mathfrak{g}$ generate an infinitesimal action $\delta_\xi \theta$. 
Then, tangent directions (spanning the tangent space to a group orbit) have the form
\begin{equation}
\delta\theta_{\parallel} \in T_\theta([\theta]) = \{ \delta_\xi \theta : \xi \in \mathfrak{g}\}.
\end{equation}
We may decompose any parameter update $\Delta \theta$ (that is, a finite optimization update, e.g., gradient step) into tangent and transverse components,
\begin{equation}
\Delta\theta = \Delta\theta_{\parallel} + \Delta\theta_{\perp},
\end{equation}
where $\Delta\theta_{\perp}$ moves in functionally meaningful directions (changing $[\theta]$) and $\Delta\theta_{\parallel}$ moves along a redundant orbit.
{Here $\Delta\theta_{\parallel}\in \mathrm{span}\{\delta_\xi\theta:\xi\in\mathfrak g\}$, i.e. $\Delta\theta_{\parallel}=\sum_a c_a\,\delta_{\xi_a}\theta$,
for some basis $\{\xi_a\}$ of $\mathfrak g$ and coefficients $c_a\in\mathbb R$.}

In ideal learning dynamics, we would enforce
\begin{equation}
\Delta\theta_{\parallel} = 0,
\end{equation}
so that learning updates are purely transverse.
Note that this is similar, morally, to what is done in gauge theory in physics, when \emph{fixing a gauge}.
\emph{Quotient-based} learning dynamics would avoid this by evolving directly in $\Theta_{\text{red}}$.
This is the spirit of the DFM.

\paragraph{A DFM-inspired dressing map for optimization.\\}

In physics, in gauge theory, symmetry reduction can be achieved by constructing dressed variables via a dressing field that removes the redundant group action. Translating this logic to optimization suggests introducing a projection (surjective) map
\begin{equation}
\begin{aligned}
 \mathcal{D}: \Theta &\to \Theta_{\mathrm{red}}, \\
 \theta &\mapsto \tilde{\theta}:=\mathcal{D}(\theta) = \mathcal{D}(g\cdot \theta).
\end{aligned}
\end{equation}
that assigns to each $\theta$ a representative in the reduced space, or equivalently a set of invariant coordinates. 
Optimization then proceeds by updating $\tilde{\theta}$ and, when needed, reconstructing a compatible $\theta$ (a representative) for forward computation. Geometrically, we have the \emph{pushforward map} $\mathcal D_*: T_\theta \Theta \rarrow \Theta_{\mathrm{red}}$ acting as
\begin{equation}
\label{Red-dynamics}
\begin{aligned}
 \mathcal{D}_*(\Delta \theta_\parallel) =0
 \quad \text{so} \quad
 \mathcal{D}_*(\Delta \theta)
 =\mathcal{D}_*(\Delta \theta_\perp)
 \in T_{\tilde \theta} \Theta_{\mathrm{red}}.
\end{aligned}
\end{equation}
Thus the optimal update is $\Delta \tilde \theta:=\mathcal{D}_*(\Delta \theta_\perp)$.

Observe that 
this differs from explicit symmetry breaking: no preferred direction is introduced; rather, the redundant d.o.f. never appear as independent optimization variables.

\subsubsection{Reduced variables as invariants: Learning on $\Theta_{\mathrm{red}}$}

A practical approach to quotient optimization is to parameterize the model using \emph{invariants} under the group $G$. 
In the attention setting, an example of such invariant combinations arises from the observation that dot-product scores depend on $W_Q$ and $W_K$ only through the bilinear form
\begin{equation}
G_{QK} := W_Q^\top W_K,
\end{equation}
since
\begin{equation}
(W_Q x_i)^\top (W_K x_j) = x_i^\top (W_Q^\top W_K)\, x_j = x_i^\top G_{QK}\,x_j.
\end{equation}
Under a \emph{left} action by an orthogonal matrix $S\in O(d_h)$,
\begin{equation}
W_Q \mapsto S W_Q,\qquad W_K \mapsto S W_K,
\end{equation}
the composite
\begin{equation}
G_{QK}:=W_Q^\top W_K\in\RR^{d\times d}
\end{equation}
is unchanged since $(SW_Q)^\top(SW_K)=W_Q^\top W_K$.
Hence, $(W_Q,W_K)$ are identifiable only up to a shared $O(d_h)$ rotation of the head space.
In physics, this would correspond to the alternative move of \emph{dressing}, rather than \emph{gauge-fixing} \cite{Berghofer:2024plf,Francois:2024rdm}.

Note that,
if $d_h<d$, then $\mathrm{rank}(G_{QK})\le d_h$, so not every $d\times d$ matrix is representable as $W_Q^\top W_K$.
Thus, optimizing directly in $G_{QK}$ either requires enforcing $\mathrm{rank}(G_{QK})\le d_h$ (a constrained problem), or else allowing arbitrary $G_{QK}$, which changes the model class.

\paragraph{Value-output sector.\\}

Let $W_V\in\RR^{d_h\times d}$ and $W_O\in\RR^{d\times d_h}$ denote the value and output projections of a single head (up to architectural conventions).
The contribution of this head to the hidden update is
\begin{equation}
x_i \mapsto W_O W_V x_i.
\end{equation}
Define the \emph{composite} (which can be seen as a dressed quantity)
\begin{equation}
G_{VO}:=W_O W_V\in\RR^{d\times d}.
\end{equation}
Under a right action by any invertible matrix $S\in GL(d_h,\RR)$,
\begin{equation}
W_V\mapsto S^{-1}W_V,\qquad W_O\mapsto W_O S,
\end{equation}
the composite $G_{VO}$ is unchanged:
\begin{equation}
(W_O S)(S^{-1}W_V)=W_O W_V.
\end{equation}
Hence, $(W_O,W_V)$ are identifiable only up to this internal $GL(d_h,\RR)$ reparameterization.

\smallskip

Since $G_{VO}$ factors through a $d_h$-dimensional latent space, it necessarily satisfies
\begin{equation}
\mathrm{rank}(G_{VO})\le d_h.
\end{equation}
Consequently, optimizing directly over $G_{VO}$ either requires enforcing the rank constraint, or, if arbitrary $d\times d$ matrices are allowed, amounts to optimizing over a different model family.

A simple projected gradient step in the value-output sector is sketched in Appendix \ref{Algorithmic sketch: Invariant updates for the value-output sector}.

\paragraph{Multi-head symmetry and head permutations.\\}

For $H$ attention heads, the forward computation is invariant under arbitrary permutations of the heads.
Let $\sigma\in S_H$ act by permuting head indices simultaneously in $(W_Q^{(h)},W_K^{(h)},W_V^{(h)},W_O^{(h)})$.
Then the summed output over heads is unchanged.
Thus, in addition to the continuous reparameterizations described above, there is a discrete symmetry group $S_H$ acting on parameters.

Functionally distinct models therefore live on the quotient space
\begin{equation}
\Theta_{\mathrm{red}}
=\Theta/\Big(O(d_h)_{\text{QK}}^H\times GL(d_h,\RR)_{\text{VO}}^H\times S_H\Big),
\end{equation}
up to architectural details.

\smallskip

Note that the discrete permutation symmetry $S_H$ differs conceptually from the continuous reparameterizations discussed above. 
While the groups $O(d_h)^H$ and $GL(d_h,\RR)^H$ generate infinitesimal tangent directions in parameter space and are responsible for genuine optimization degeneracies, the action of $S_H$ is purely \emph{discrete} and corresponds only to a relabeling of identical attention heads.
$S_H$ induces no flat directions, conserved quantities, or continuous orbit structure.
As such, one may regard it as a \emph{residual reference-frame freedom}.

\paragraph{Projected gradient flow.\\}

Let us mention that a complementary approach is to keep the original parameterization but project updates to the transverse subspace. Let $\nabla \mathcal{L}(\theta)$ denote the gradient in $\Theta$ equipped with an inner product $\langle\cdot,\cdot\rangle$. 
Let $P_\perp(\theta)$ be the orthogonal projector onto the transverse complement of $T_\theta([\theta])$. 
Here orthogonality is defined with respect to a chosen Riemannian metric on parameter space.
This choice is not canonical and affects the reduced dynamics.
In practice, exact projection is expensive, so symmetry reduction is typically implemented in structured approximations (e.g., per-head reductions where tangent directions are known analytically).

Then, the projected gradient flow is
\begin{equation}
\dot{\theta} = - P_\perp(\theta)\,\nabla \mathcal{L}(\theta).
\end{equation}
Discretizations of this flow yield reduced versions of Stochastic Gradient Descent (SGD) or momentum methods.
{This quotient-based projection viewpoint is conceptually related to projection-based gradient modification methods such as PCGrad \cite{Yu2020PCGrad}, although the objective here is different: we project out symmetry-tangent directions associated with internal reparameterization redundancies, rather than resolving conflicts among gradients coming from multiple tasks.}

From this perspective, standard optimizers can be interpreted as evolving on $\Theta$ with implicit (often uncontrolled) components in both tangent and transverse directions, whereas a projected optimizer explicitly removes the tangent component.
Notice the difference with \eqref{Red-dynamics}.

{Note that, as in other projection-based optimization schemes, removing selected update components may improve conditioning in some regimes but may also increase per-step cost or alter beneficial stochastic effects.
Hence, whether quotient-based projection is advantageous is task- and implementation-dependent, and should be assessed empirically.}

\paragraph{Reduced Hamiltonian dynamics and conserved quantities.\\}

When optimization is formulated dynamically -- for instance in Hamiltonian or momentum-based forms -- symmetries can induce conserved quantities. 
Consider a generic Hamiltonian-like system on $(\theta,\pi)$:
\begin{equation}
\dot{\theta} = \frac{\partial H}{\partial \pi},
\qquad
\dot{\pi} = -\frac{\partial H}{\partial \theta},
\end{equation}
where $\pi$ denotes momenta. If $H$ is invariant under a continuous group action, then the associated Noether charges are conserved in idealized dynamics, constraining exploration in phase space.

A symmetry-reduced formulation replaces the full phase space by a reduced phase space:
\begin{equation}
(\Theta\times \Pi)/G,
\end{equation}
in which the Noether charges are absent as independent d.o.f.; Equivalently, the reduced system evolves only in the functionally relevant directions, eliminating conserved motions along redundant symmetry orbits.

This makes clear why symmetry reduction is a principled alternative to symmetry breaking: rather than introducing stochastic perturbations or preferred directions to shake the system out of conserved motions, we can remove the conserved directions by construction.

\subsection{Practical schemes for symmetry-free optimization}

Let us therefore outline three practical schemes for optimizing in reduced spaces:
\begin{enumerate}
    \item[(i)] \emph{Dressed representatives.} 
    Maintain a representative $\theta$ for forward computation while regularly applying a dressing map $\mathcal{D}$ that removes accumulated motion along the orbit. 
    \item[(ii)] \emph{Invariant reparameterization.} 
    Replace redundant parameters by invariant composites (e.g., $G_{QK}$, $G_{VO}$) and train directly in these variables. This eliminates redundant motions but may change computational cost and requires ensuring sufficient expressivity and numerical stability.
    \item[(iii)] \emph{Projection-based quotient updates.} 
    Maintain standard parameters but project gradients (or momentum) orthogonally to symmetry orbits. This requires computing tangent directions and implementing projectors, but keeps the forward pass unchanged.
\end{enumerate}
Each scheme implements the same principle: learning updates should occur only in the reduced space of functionally distinct models.

\smallskip

This section formalized optimization as a dynamical process on a parameter space with symmetry-induced redundancies. 
We proposed a symmetry-reduced viewpoint in which training takes place on a quotient space $\Theta/G$, either by reparameterizing in invariant variables or by explicitly projecting updates to remove tangent components along symmetry orbits.

In summary, attention layers exhibit continuous internal symmetries $O(d_h)$ (query-key), $GL(d_h,\RR)$ (value-output), and a discrete head permutation symmetry $S_H$.
The functionally meaningful d.o.f. are therefore encoded in invariant composites such as $G_{QK}$ and $G_{VO}$ subject to rank constraints, together with their arrangement across heads.
Symmetry-reduced optimization amounts to learning directly on this quotient space rather than on redundant parameterizations.

{For clarity, we now provide a minimal concrete differentiable realization and initialization strategy for a representative reduced sector.}

\subsection{{Concrete differentiable realization and initialization sketch}}
\label{Concrete realization and initialization sketch}

{To clarify how symmetry-reduced variables may be handled in practice, we describe a minimal differentiable realization for a representative case, namely the value-output sector of a single attention head. 
This sector contributes to the hidden update through the composite operator
$G_{VO} := W_O W_V \in \RR^{d\times d}$,
which is invariant under the internal reparameterization
$(W_V,W_O) \mapsto (S^{-1}W_V,W_O S)$, with $S\in GL(d_h,\RR)$.}

{A convenient realization of this invariant is obtained by introducing matrices $A \in \RR^{d\times d_h}$, $B \in \RR^{d_h\times d}$,
and defining $G_{VO} = AB$,
which enforces $\mathrm{rank}(G_{VO}) \le d_h$ by construction.
The pair $(A,B)$ provides a representation of the invariant operator $G_{VO}$, and it is therefore not unique. 
In particular, it admits a residual redundancy
$(A,B) \mapsto (A S, S^{-1} B)$, with $S \in GL(d_h,\RR)$,
which leaves the composite $G_{VO}$ unchanged and pertains only to the chosen representation of $G_{VO}$.}

{Now, as far as differentiable implementation is concerned, we observe that the map $(A,B)\mapsto AB$ is smooth, so gradients of the loss with respect to $A$ and $B$ can be computed by standard automatic differentiation through the matrix product. 
No modification of backpropagation algorithms is required.}

{A natural initialization is then obtained by sampling $A$ and $B$ using standard variance-preserving random initializations (e.g., Xavier/Glorot-type schemes adapted to their dimensions), and setting $G_{VO}^{(0)} = A_0 B_0$.}

{Alternatively, one may treat $G_{VO}$ itself as the optimization variable in $\RR^{d\times d}$, perform gradient updates directly on $G_{VO}$, and enforce the rank constraint $\mathrm{rank}(G_{VO}) \le d_h$ by projection (for instance via truncated SVD), as described in Appendix \ref{Algorithmic sketch: Invariant updates for the value-output sector}. An initial $G_{VO}^{(0)}$ may again be obtained from a low-rank factorization.}

\smallskip

{This construction provides a concrete differentiable realization for a representative symmetry-reduced sector. 
It illustrates how invariant variables, gradients, and initialization can be handled within standard automatic-differentiation-based machine learning frameworks, yet without specifying a complete training procedure for full transformer architectures.}

\section{Conclusions}
\label{Conclusions}

We outline a conceptual framework for relational, symmetry-reduced transformer architectures, in which:
\begin{enumerate}
\item Internal representation frames are dressed by replacing coordinate-dependent vectors with relational invariants, and attention mechanisms are rewritten in invariant relational form.
\item Optimization dynamics are studied on reduced parameter spaces, eliminating motion along redundant symmetry orbits.
\end{enumerate}
Together, these would yield a framework in which representations, attention, optimization, and structure are all formulated in manifestly relational (with invariant weights and equivariant vector carriers).

Of course, both standard dot-product attention and explicit relational matrices involve pairwise token interactions and therefore {retain an $O(n^2)$ interaction structure in sequence length.}
Standard implementations propagate per-token vectors and compute pairwise scores on the fly, whereas a fully relational formulation may choose to store relational objects across layers.
Storing relational objects across depth can increase memory unless one uses low-rank/sparse parameterizations or recomputation strategies.

{Empirical evaluation of these ideas is left for future work (possibly collaborative).
In particular, whether symmetry reduction yields measurable gains in wall-clock training time, memory consumption, or convergence speed relative to standard transformer implementations is an open implementation-dependent question that requires dedicated benchmarking on shared tasks and matched model budgets.
The present paper provides the conceptual and mathematical framework needed to formulate such comparisons precisely, but does not itself report experimental results.}

{For future practical implementations, beyond predictive accuracy one should also track resource and optimization metrics such as:}
\begin{itemize}
    \item Training loss curves (does the invariant version converge faster or more stably?),
    \item Sensitivity to random initialization (variance across 20-50 seeds),
    \item Effective rank of internal representations or parameter matrices (does symmetry reduction lead to lower effective dimensionality?),
    \item Qualitative attention patterns (are they more interpretable in the relational case?).
\end{itemize}

Future directions include developing scalable implementations of invariant relational attention and designing optimizers that operate directly on reduced parameter spaces.
{We also note that introducing architectural inductive biases -- for instance, through locality, hierarchy, or constrained attention patterns -- is complementary in spirit to the present proposal.
Such biases may improve optimization and generalization for reasons different from symmetry reduction, and studying their interplay with the reduced relational variables proposed here would be an interesting direction for future work.}

More broadly, symmetry reduction provides a unifying lens for understanding representation and learning dynamics in modern ML. 
By removing redundant d.o.f., we move toward models whose internal structure directly reflects the relational organization of the data they process.
We hope this framework opens new avenues for both theoretical analysis and practical architecture design in deep learning.

\subsection{{Empirical scope and evaluation roadmap}}
\label{Empirical scope and evaluation roadmap}

{A natural question is whether the symmetry-reduced formulations proposed here lead, in practice, to improved training efficiency, lower memory usage, or better predictive performance relative to standard transformer baselines.
We emphasize that the present paper does not answer this experimentally.
Its objective is to isolate the relevant symmetries, formulate invariant relational variables, and identify reduced parameterizations and projected dynamics at the conceptual and mathematical level.}

{Nevertheless, the framework suggests a concrete experimental program.
A minimal empirical study would compare:
(i) a standard transformer baseline,
(ii) a relational/invariant scoring variant in model space, and/or
(iii) a symmetry-reduced optimization variant in head space,
on the same task and with matched parameter budgets as closely as possible.
Relevant measurements would include:
\begin{enumerate}
    \item predictive performance (validation/test loss or accuracy),
    \item wall-clock training time per epoch and to target loss,
    \item peak memory consumption,
    \item number of trainable parameters and effective rank of learned operators,
    \item sensitivity to initialization across multiple random seeds,
    \item optimizer statistics indicating motion along reduced \emph{vs} redundant directions when projection-based schemes are used.
\end{enumerate}}

{From a computational viewpoint, both standard dot-product attention and the relational scoring constructions discussed in this paper involve pairwise token interactions and therefore share the same $O(n^2)$ interaction structure in sequence length.
However, their constant factors, memory access patterns, and implementation overheads may differ substantially.
In particular, explicitly storing relational objects such as $G=XX^\top$ across layers may increase memory usage unless low-rank, sparse, kernelized, or recomputation strategies are adopted.
Likewise, quotient- or projection-based optimization may reduce motion along redundant parameter directions, but whether this translates into lower wall-clock cost or improved convergence is an empirical question.}

{Accordingly, the present work should be read as providing the mathematical framework and architectural principles that make such an experimental comparison well-posed, rather than as reporting the outcome of that comparison.
A full empirical evaluation is left for future work, potentially in collaboration with researchers specializing in large-scale implementation and benchmarking.
A possible starting point for such future benchmarking is to follow the experimental paradigm of \cite{silverstein:2026pdf}, adapting it to compare standard transformer baselines with the symmetry-reduced variants proposed here under matched parameter and training budgets.}

\begin{acknowledgments}
J.F. is supported by the Austrian Science Fund (FWF), \mbox{[P 36542]} and by the Czech Science Foundation (GAČR), grant GA24-10887S.
L.R. is supported by the research grant PNRR Young Researchers, funded by MUR, MSCA Seal of Excellence (SoE), CUP E13C24003600006, ID SOE2024$\_$0000103, project GrIFOS, of which this paper is part.
\end{acknowledgments}

\section*{Author contribution declaration}

Both authors share equal credit for the work done in this paper: J.F. and L.R. conceived of the presented idea, J.F. and L.R. developed the formalism. Both J.F. and L.R. contributed equally to the writing of the manuscript.

\section*{Funding declaration}

The research was supported by the Austrian Science Fund (FWF), grant \mbox{P 36542}, the Czech Science Foundation (GAČR), grant GA24-10887S, and the GrIFOS research project, funded by MUR, Italy, CUP E13C24003600006, ID SOE2024$\_$0000103.

\section*{Data availability statement}

No data were created or analyzed in this study.

\appendix

\section{Algorithmic sketch: Invariant updates for the value-output sector}\label{Algorithmic sketch: Invariant updates for the value-output sector}

This appendix sketches an invariant (symmetry-reduced) optimization viewpoint for the value-output parameters
$(W_V \in \RR^{d_h \times d},\, W_O \in \RR^{d \times d_h})$
of a single attention head.
As discussed in Section \ref{Optimization dynamics on reduced, symmetry-free parameter spaces}, the forward contribution of this sector depends on $(W_V,W_O)$ only through the composite
\begin{equation}
G_{VO} := W_O W_V \in \RR^{d\times d},
\end{equation}
which is invariant under the internal reparameterization
\begin{equation}
(W_V,W_O)\mapsto(S^{-1}W_V,\;W_O S),
\quad S\in GL(d_h,\RR).
\end{equation}
In the idealized attention submodule (i.e. in the absence of symmetry-breaking components such as LayerNorm or nonlinear MLPs), this composite therefore captures the functionally relevant d.o.f. of the value-output sector.

From the symmetry-reduction perspective advocated in this paper, it is natural to regard $G_{VO}$ as the primary variable and to formulate learning directly in terms of this invariant.
{Here let us also remark that, in practice, gradients with respect to the invariant composites can be obtained by standard automatic differentiation once a concrete differentiable realization of the reduced variables is chosen.}

\subsection{Invariant gradient and direct update on $G_{VO}$}

Restricting attention to the symmetric submodule, the loss may be viewed as a function of the invariant,
\begin{equation}
\mathcal{L}=\mathcal{L}(G_{VO}),
\end{equation}
up to symmetry-breaking components such as LayerNorm, biases, and nonlinear MLPs (cf. Section \ref{What symmetry actually holds (architecture-dependent)}).
One may therefore define the invariant gradient
\begin{equation}
\nabla_{G_{VO}}\mathcal{L}\;\in\;\RR^{d\times d},
\end{equation}
and perform symmetry-reduced gradient descent directly on $G_{VO}$:
\begin{equation}
G_{VO} \;\leftarrow\; G_{VO} - \eta \nabla_{G_{VO}}\mathcal{L},
\label{eq:GVO_update}
\end{equation}
with learning rate \(\eta>0\) ($\eta \in \RR_{>0}$).
This update evolves on the reduced parameter space and contains no motion along the internal
\(GL(d_h,\RR)\) symmetry directions, since these do not appear as independent variables.

\subsection{Rank constraint and low-rank realizations}
\label{Rank constraint and low-rank realizations}

Since $G_{VO}$ factors through a $d_h$-dimensional head space, it necessarily satisfies
\begin{equation}
\text{rank}(G_{VO}) \le d_h.
\end{equation}
Consequently, unrestricted updates of the form \eqref{eq:GVO_update} may leave the representable set unless this constraint is enforced (if one updated $G_{VO}$ freely, it may become full-rank, hence no longer realizable by any $(W_O,W_V)$).
Two standard strategies may be used to address this.

\begin{enumerate}
\item[(i)] \emph{Explicit low-rank parameterization}: 
One may parameterize the invariant directly as
\begin{equation}
G_{VO} = A B,
\quad
A\in\RR^{d\times d_h},\;\; B\in\RR^{d_h\times d},
\end{equation}
which enforces $\text{rank}(G_{VO}) \le d_h$ by construction.
Gradient-based optimization is then carried out on $(A,B)$, with the model depending on these variables only through their product $AB$.
Note that $A$, $B$ merely provide a low-rank realization of the invariant operator.
The factorization itself retains a residual internal reparameterization freedom
$(A,B)\mapsto(AS,S^{-1}B)$, $S\in GL(d_h,\RR)$,
but this redundancy pertains only to the chosen representation of $G_{VO}$; the learned function depends solely on the invariant composite.

\item[(ii)] \emph{Projected invariant updates}:
Alternatively, one may update $G_{VO}$ in the ambient space via \eqref{eq:GVO_update} and subsequently project back onto the rank-$\le d_h$ manifold.
A simple projection is obtained via (rank-$\le d_h$) truncated SVD:
\begin{equation}
\begin{aligned}
& G_{VO} = U \Sigma V^\top, \\
& \Pi_{\le d_h}(G_{VO})
:= U_{d_h}\,\Sigma_{d_h}\,V_{d_h}^\top,
\end{aligned}
\end{equation}
where $G_{VO}=U\Sigma V^\top$ is an SVD with singular values in descending order, 
$U_{d_h}\in\RR^{d\times d_h}$ and $V_{d_h}\in\RR^{d\times d_h}$ denote the matrices formed by the first $d_h$ columns of $U$ and $V$, respectively, and
$\Sigma_{d_h}\in\RR^{d_h\times d_h}$ is the diagonal matrix of the largest $d_h$ singular values.
{The SVD factorizes $G_{VO}$ into orthogonal matrices $U$, $V$ and a diagonal matrix $\Sigma$ of nonnegative singular values, providing the optimal rank-$\le d_h$ approximation in Frobenius norm (the truncated SVD gives the lowest possible total squared error when compressing $G_{VO}$ to rank-$\le d_h$).}

The invariant update then reads
\begin{equation}
G_{VO} \;\leftarrow\; \Pi_{\le d_h}\!\left(G_{VO}-\eta \nabla_{G_{VO}}\mathcal{L}\right).
\end{equation}
In other words, one first performs a gradient step in the ambient space of $d\times d$ matrices and then projects back onto the rank-$\le d_h$ manifold to ensure compatibility with the attention-head structure.
\end{enumerate}

\subsection{Choosing representatives for standard implementations}

If one wishes to implement the forward pass in the conventional factorized form $W_O$, $W_V$, a representative factorization of the current invariant $G_{VO}$ may be chosen at any stage.
For example, from a truncated SVD $G_{VO}=U\Sigma V^\top$, with $\Sigma\in\RR^{d_h\times d_h}$, one may set
\begin{equation}
W_O := U \Sigma^{1/2},
\quad
W_V := \Sigma^{1/2} V^\top,
\end{equation}
so that $W_O W_V = G_{VO}$.
This choice is not unique: any pair $(W_O S, S^{-1}W_V)$, with $S\in GL(d_h,\RR)$, yields the same invariant.
Of course, such choices correspond merely to different internal coordinate frames for the same relational operator.

\smallskip

In conclusion, in this symmetry-reduced formulation, the value-output sector is naturally described by the invariant composite $G_{VO}=W_O W_V$.
Learning may therefore proceed directly on this object, with rank constraints enforced either by low-rank parameterization or by projection.
This realizes symmetry reduction by construction: optimization evolves only in functionally meaningful directions, without introducing preferred internal frames or explicit symmetry breaking.

\bibliography{biblio.bib}

@book{GuilleminSternberg1990,
  author    = {Guillemin, Victor and Sternberg, Shlomo},
  title     = {Symplectic Techniques in Physics},
  publisher = {Cambridge University Press},
  year      = {1990},
  address   = {Cambridge},
  isbn      = {978-0521389907},
  note      = {Paperback reprint (original hardcover 1984)}
}

@book{HenneauxTeitelboim1992,
  author    = {Henneaux, Marc and Teitelboim, Claudio},
  title     = {Quantization of Gauge Systems},
  publisher = {Princeton University Press},
  year      = {1992},
  address   = {Princeton, NJ},
  isbn      = {978-0691037691},
  note      = {Paperback edition (original hardcover also 069108775X)}
}

@article{Gieres2023,
  author    = {Gieres, Fran{\c{c}}ois},
  title     = {Covariant canonical formulations of classical field theories},
  journal   = {SciPost Physics Lecture Notes},
  number    = {77},
  year      = {2023},
  month     = {dec},
  publisher = {SciPost Foundation},
  doi       = {10.21468/SciPostPhysLectNotes.77},
  note      = {arXiv:2109.07330},
}

@article{Francois:2024vlr,
    author = "Fran{\c{c}}ois, J. and Ravera, L.",
    title = "{On the Meaning of Local Symmetries: Epistemic-ontological Dialectics}",
    eprint = "2404.17449",
    archivePrefix = "arXiv",
    primaryClass = "physics.hist-ph",
    doi = "10.1007/s10701-025-00849-y",
    journal = "Found. Phys.",
    volume = "55",
    number = "3",
    pages = "38",
    year = "2025"
}

@article{Francois:2024laf,
    author = "Fran{\c{c}}ois, J. and Ravera, L.",
    title = "{Dressing fields for supersymmetry: the cases of the Rarita-Schwinger and gravitino fields}",
    eprint = "2405.04379",
    archivePrefix = "arXiv",
    primaryClass = "hep-th",
    doi = "10.1007/JHEP07(2024)041",
    journal = "JHEP",
    volume = "07",
    pages = "041",
    year = "2024"
}

@article{Francois:2024rdm,
    author = "Fran{\c{c}}ois, J. T. and Ravera, L.",
    title = "{Geometric Relational Framework for General-Relativistic Gauge Field Theories}",
    eprint = "2407.04043",
    archivePrefix = "arXiv",
    primaryClass = "gr-qc",
    doi = "10.1002/prop.202400149",
    journal = "Fortsch. Phys.",
    volume = "73",
    number = "1-2",
    pages = "2400149",
    year = "2025"
}

@article{Francois:2024xqi,
    author = "Fran{\c{c}}ois, J. and Ravera, L.",
    title = "{Unconventional supersymmetry via the dressing field method}",
    eprint = "2412.01898",
    archivePrefix = "arXiv",
    primaryClass = "hep-th",
    doi = "10.1103/76n5-4mg1",
    journal = "Phys. Rev. D",
    volume = "111",
    number = "12",
    pages = "125022",
    year = "2025"
}

@article{Francois:2025shu,
    author = "Fran{\c{c}}ois, J. T. and Ravera, L.",
    title = "{Relational Bundle Geometric Formulation of Non-Relativistic Quantum Mechanics}",
    eprint = "2501.02046",
    archivePrefix = "arXiv",
    primaryClass = "quant-ph",
    doi = "10.1002/prop.70040",
    journal = "Fortsch. Phys.",
    volume = "73",
    number = "12",
    pages = "e70040",
    year = "2025"
}

@article{Francois:2025odk,
    author = "Fran{\c{c}}ois, J. and Ravera, L.",
    title = "{Off-shell supersymmetry via manifest invariance}",
    eprint = "2504.06392",
    archivePrefix = "arXiv",
    primaryClass = "hep-th",
    doi = "10.1016/j.physletb.2025.139633",
    journal = "Phys. Lett. B",
    volume = "868",
    pages = "139633",
    year = "2025"
}

@article{Francois:2025sic,
    author = "Fran{\c{c}}ois, J. and Ravera, L.",
    title = "{Spacetime boundaries do not break diffeomorphism and gauge symmetries}",
    eprint = "2504.20945",
    archivePrefix = "arXiv",
    primaryClass = "gr-qc",
    doi = "10.1103/pwv6-tg7n",
    journal = "Phys. Rev. D",
    volume = "112",
    number = "12",
    pages = "125029",
    year = "2025"
}

@article{Francois:2025jro,
    author = "Fran{\c{c}}ois, J. and Ravera, L.",
    title = "{Reassessing the foundations of metric-affine gravity}",
    eprint = "2505.05349",
    archivePrefix = "arXiv",
    primaryClass = "gr-qc",
    doi = "10.1140/epjc/s10052-025-14656-2",
    journal = "Eur. Phys. J. C",
    volume = "85",
    pages = "902",
    year = "2025"
}

@article{Berghofer:2025ius,
    author = "Berghofer, P. and Fran{\c{c}}ois, J. and Ravera, L.",
    title = "{What Price Fiber Bundle Substantivalism? On How to Avoid Holes in Fibers}",
    journal = "arXiv:2505.12876 [physics.hist-ph]",
    year = "2025"
}

@article{Francois:2025lqn,
    author = "Fran{\c{c}}ois, J. and Ravera, L.",
    title = "{Mechanics as a general-relativistic gauge field theory, and Relational Quantization}",
    journal = "arXiv:2510.19845 [physics.gen-ph]",
    year = "2025"
}

@article{Francois:2025ptj,
    author = "Fran{\c{c}}ois, J. and Ravera, L.",
    title = "{Raising galaxy rotation curves via dressing}",
    eprint = "2510.18549",
    archivePrefix = "arXiv",
    primaryClass = "gr-qc",
    doi = "10.1103/m9xl-9vvk",
    journal = "Phys. Rev. D",
    volume = "112",
    pages = "L081501",
    year = "2025"
}

@article{Berghofer:2024plf,
    author = "Berghofer, P. and Fran{\c{c}}ois, J.",
    title = "{Dressing vs. Fixing: On How to Extract and Interpret Gauge-Invariant Content}",
    eprint = "2404.18582",
    archivePrefix = "arXiv",
    primaryClass = "physics.hist-ph",
    doi = "10.1007/s10701-024-00809-y",
    journal = "Found. Phys.",
    volume = "54",
    number = "6",
    pages = "72",
    year = "2024"
}

@article{FrancoisAndre:2023jmj,
    author = "Francois André, J. T.",
    title = "{The dressing field method for diffeomorphisms: a relational framework}",
    eprint = "2310.14472",
    archivePrefix = "arXiv",
    primaryClass = "math-ph",
    doi = "10.1088/1751-8121/ad5cad",
    journal = "J. Phys. A",
    volume = "57",
    number = "30",
    pages = "305203",
    year = "2024"
}

@article{Francois:2017akk,
    author = "Fran{\c{c}}ois, J.",
    title = "{Artificial versus Substantial Gauge Symmetries: A Criterion and an Application to the Electroweak Model}",
    eprint = "1801.00678",
    archivePrefix = "arXiv",
    primaryClass = "physics.hist-ph",
    doi = "10.1086/703571",
    journal = "Phil. Sci.",
    volume = "86",
    number = "3",
    pages = "472--496",
    year = "2019"
}

@article{silverstein:2026pdf,
    title={Symmetry Breaking in Transformers for Efficient and Interpretable Training},
    author={Silverstein, E. and Kunin, D. and Shyam, V.},
    journal={arXiv:2601.22257},
    year={2026}
}

@article{vaswani2017attention,
  title={Attention Is All You Need},
  author={Vaswani, Ashish and Shazeer, Noam and Parmar, Niki and Uszkoreit, Jakob and Jones, Llion and Gomez, Aidan N. and Kaiser, Lukasz and Polosukhin, Illia},
  booktitle={Advances in Neural Information Processing Systems (NeurIPS)},
  journal={arXiv:1706.03762},
  year={2017}
}

@article{devlin2018bert,
  title={BERT: Pre-training of Deep Bidirectional Transformers for Language Understanding},
  author={Devlin, Jacob and Chang, Ming-Wei and Lee, Kenton and Toutanova, Kristina},
  booktitle={NAACL},
  journal={arXiv:1810.04805},
  year={2019}
}

@article{brown2020gpt3,
  title={Language Models are Few-Shot Learners},
  author={Brown, Tom B. and Mann, Benjamin and Ryder, Nick and Subbiah, Melanie and Kaplan, Jared et al},
  booktitle={NeurIPS},
  journal={arXiv:2005.14165},
  year={2020}
}

@article{dosovitskiy2020vit,
  title={An Image is Worth 16x16 Words: Transformers for Image Recognition at Scale},
  author={Dosovitskiy, Alexey and Beyer, Lucas and Kolesnikov, Alexander et al},
  booktitle={ICLR},
  journal={arXiv:2010.11929},
  year={2021}
}

@article{radford2021clip,
  title={Learning Transferable Visual Models From Natural Language Supervision},
  author={Radford, Alec and Kim, Jong Wook et al},
  booktitle={ICML},
  journal={arXiv:2103.00020},
  year={2021}
}

@article{kunin2020neuralmechanics,
  title={Neural Mechanics: Symmetry and Broken Conservation Laws in Deep Learning Dynamics},
  author={Kunin, Daniel and Sagastuy-Brena, Javier and Ganguli, Surya and Yamins, Daniel L. K. and Tanaka, Hidenori},
  journal={arXiv:2012.04728},
  year={2020}
}

@article{tanaka2021noether,
  title={Noether’s Learning Dynamics: Role of Symmetry Breaking in Neural Networks},
  author={Tanaka, Hidenori and Kunin, Daniel},
  journal={arXiv:2105.02716},
  year={2021}
}

@article{zhang2025rotationsymmetry,
  title={Beyond the Permutation Symmetry of Transformers: The Role of Rotation for Model Fusion},
  author={Zhang, Binchi and Zheng, Zaiyi and Chen, Zhengzhang and Li, Jundong},
  journal={arXiv:2502.00264},
  year={2025}
}

@article{battaglia2018relational,
  title={Relational Inductive Biases, Deep Learning, and Graph Networks},
  author={Battaglia, Peter W. and Hamrick, Jessica B. et al},
  journal={arXiv:1806.01261},
  year={2018}
}

@article{santoro2017relationnetworks,
  title={A Simple Neural Network Module for Relational Reasoning},
  author={Santoro, Adam and Raposo, David and Barrett, David G.T. and Malinowski, Mateusz and Pascanu, Razvan and Battaglia, Peter and  Lillicrap, Timothy},
  booktitle={NeurIPS},
  journal={arXiv:1706.01427},
  year={2017}
}

@article{zaheer2017deepsets,
  title={Deep Sets},
  author={Zaheer, Manzil and Kottur, Satwik and Ravanbakhsh, Siamak and Poczos, Barnabas and Salakhutdinov, Ruslan and Smola, Alexander},
  journal={arXiv:1703.06114},
  year={2017}
}

@article{zhou2023survey,
  title={A Survey of Transformers},
  author={Lian, Tianyang and Wang, Yuxin and Liu, Xiangyang and Qiu, Xipeng},
  journal={arXiv:2106.04554},
  year={2023}
}

@article{elhage2023circuits,
  title={A Mathematical Framework for Transformer Circuits},
  author={Elhage, Nelson and Nanda, Neel and Olsson, Catherine and Henighan, Tom and Joseph, Nicholas and Mann, Ben and Askell, Amanda and Bai, Yuntao and Chen, Anna and Conerly, Tom et al},
  journal={ https://transformer-circuits.pub/2021/framework/index.html},
  year={2023}
}

@article{cohen2016gcnn,
  title={Group Equivariant Convolutional Networks},
  author={Cohen, Taco S. and Welling, Max},
  booktitle={International Conference on Machine Learning (ICML)},
  journal={arXiv:1602.07576},
  year={2016}
}

@article{kondor2018generalization,
  title={On the Generalization of Equivariance and Convolution in Neural Networks to the Action of Compact Groups},
  author={Kondor, Risi and Trivedi, Shubhendu},
  booktitle={International Conference on Machine Learning (ICML)},
  journal={arXiv:1802.03690},
  year={2018}
}

@article{merrill2020formal,
  title={A Formal Hierarchy of RNNs and Transformers},
  author={Merrill, William and Weiss, Gail and Goldberg, Yoav and Schwartz, Roy and Smith, Noah A. and Yahav, Eran},
  journal={arXiv:2004.08500},
  year={2020}
}

@article{dwivedi2020graphtransformer,
  title={A Generalization of Transformer Networks to Graphs},
  author={Dwivedi, Vijay Prakash and Bresson, Xavier},
  journal={arXiv:2012.09699},
  year={2020}
}

@inproceedings{ying2021graphormer,
  title={Do Transformers Really Perform Bad for Graph Representation?},
  author={Ying, Chengxuan and Cai, Tianle and Luo, Shengjie and Zheng, Shuxin and Ke, Guolin and He, Di and Shen, Yanming and Liu, Tie-Yan},
  booktitle={Advances in Neural Information Processing Systems (NeurIPS)},
  journal={arXiv:2106.05234},
  year={2021}
}

@article{dinh2017sharp,
  title={Sharp Minima Can Generalize For Deep Nets},
  author={Dinh, Laurent and Pascanu, Razvan and Bengio, Samy and Bengio, Yoshua},
  journal={arXiv:1703.04933},
  year={2017}
}

@article{dasilva2025hideandseek,
    title={Hide $\&$ Seek: Transformer Symmetries Obscure Sharpness $\&$ Riemannian Geometry Finds It},
    author={da Silva, Marvin F. and Dangel, Felix and Oore, Sageev},
    journal={arXiv:2505.05409},
    year={2025}
}

@article{zhao2023symmetriesconserved,
  title={Symmetries, Flat Minima and the Conserved Quantities of Gradient Flows},
  author={Zhao, Botao and Ganev, Iordan and Walters, Robin and Yu, Rose and Dehmamy, Nima},
  booktitle={International Conference on Learning Representations (ICLR)},
  journal={arXiv:2210.17216},
  year={2023},
  note={OpenReview: 9ZpciCOunFb}
}

@article{Rovelli:1990ph,
    author = "Rovelli, C.",
    title = "{What Is Observable in Classical and Quantum Gravity?}",
    reportNumber = "PITT-90-10",
    doi = "10.1088/0264-9381/8/2/011",
    journal = "Class. Quant. Grav.",
    volume = "8",
    pages = "297--316",
    year = "1991"
}

@article{Rovelli:2001bz,
    author = "Rovelli, C.",
    title = "{Partial observables}",
    eprint = "gr-qc/0110035",
    archivePrefix = "arXiv",
    doi = "10.1103/PhysRevD.65.124013",
    journal = "Phys. Rev. D",
    volume = "65",
    pages = "124013",
    year = "2002"
}

@article{Rovelli:2013fga,
    author = "Rovelli, C.",
    title = "{Why Gauge?}",
    eprint = "1308.5599",
    archivePrefix = "arXiv",
    primaryClass = "hep-th",
    doi = "10.1007/s10701-013-9768-7",
    journal = "Found. Phys.",
    volume = "44",
    number = "1",
    pages = "91--104",
    year = "2014"
}

@article{gaugeDL1,
  title={Gauge Equivariant Convolutional Networks and the Icosahedral CNN},
  author={Cohen, T. S. and Weiler, Maurice and Kicanaoglu, Berkay and Welling, Max},
  journal={arXiv:1902.04615},
  year={2019}
}

@article{theodosis2024incorporating,
  title={Incorporating gauge-invariance in equivariant networks},
  author={Theodosis, Emmanouil and Ba, Demba E. and Dehmamy, Nima},
  journal={OpenReview: YAINolpm8n},
  year={2024},
  note={OpenReview: https://openreview.net/forum?id=YAINolpm8n}
}

@article{theodosis2024constructing,
  title={Constructing gauge-invariant neural networks for scientific applications},
  author={Theodosis, Emmanouil and Ba, Demba and Dehmamy, Nima},
  journal={ICML 2024},
  year={2024}
}

@article{choi2025gauge,
  title={Gauge-Equivariant Graph Networks via Self-Interference Cancellation},
  author={Choi, Yoonhyuk and Kim, Chong-Kwon},
  journal={arXiv:2511.16062},
  year={2025}
}

@article{honda2026gauge,
  title={A Gauge-Theory-based Graph Neural Network},
  author={Honda, Hirotada},
  booktitle={Submitted to ICLR 2026},
  journal={OpenReview: QxoyccprRp},
  year={2026},
  note={OpenReview: https://openreview.net/forum?id=QxoyccprRp}
}

@article{huang2025learning,
  title={Learning Chern Numbers of Multiband Topological Insulators with Gauge Equivariant Neural Networks},
  author={Huang, Longde and Balabanov, Oleksandr and Linander, Hampus and Granath, Mats and Persson, Daniel and Gerken, Jan E.},
  booktitle={Advances in Neural Information Processing Systems (NeurIPS)},
  journal={NeurIPS 2025 Poster; arXiv:2502.15376},
  year={2025},
  note={OpenReview: https://openreview.net/forum?id=6pjzFIyFBo}
}

@article{strunk2025gauge,
  title={Gauge Flow Models},
  author={Strunk, Alexander and Assam, Roland},
  journal={arXiv:2507.13414},
  year={2025}
}

@inproceedings{Yu2020PCGrad,
  author    = {Tianhe Yu and Saurabh Kumar and Abhishek Gupta and Sergey Levine and Karol Hausman and Chelsea Finn},
  title     = {Gradient Surgery for Multi-Task Learning},
  booktitle = {Advances in Neural Information Processing Systems 33 (NeurIPS 2020)},
  year      = {2020},
  pages     = {5824--5836, arXiv:2001.06782 [cs.LG]}
}

\end{document}